\documentclass{article}

\usepackage{spconf,amsmath,graphicx}

\usepackage[utf8]{inputenc} 
\usepackage[T1]{fontenc}    
\usepackage{hyperref}       
\usepackage{url}            
\usepackage{booktabs}       
\usepackage{amsfonts}       
\usepackage{xcolor}         

\usepackage{array}
\usepackage{makecell}

\usepackage{adjustbox}
\usepackage{enumitem}
\usepackage{lipsum}
\usepackage[normalem]{ulem}

\definecolor{mypink1}{rgb}{0.858, 0.188, 0.478}

\title{Visual Representation Learning with Self-Supervised Attention for \textit{Low-Label} \textit{High-data} Regime}
\name{Prarthana Bhattacharyya*\thanks{*Work done during internship at Amazon Inc. }$\dagger$, Chenge Li, Xiaonan Zhao, Istv\'an Feh\'erv\'ari, Jason Sun}
\address{$\dagger$University of Waterloo, Amazon Inc. \\ $\dagger$p6bhatta@uwaterloo.ca, \{lichenge, xiaonzha, istvanfe, jasun\}@amazon.com}

\begin{document}
%
\maketitle
\begin{abstract}
Self-supervision has shown outstanding results for natural language processing, and more recently, for image recognition. Simultaneously, vision transformers and its variants have emerged as a promising and scalable alternative to convolutions on various computer vision tasks. 
In this paper, we are the first to question if self-supervised vision transformers (SSL-ViTs) can be adapted to two important computer vision tasks in the low-label, high-data regime:
few-shot image classification and zero-shot image retrieval. The motivation is to reduce the number of manual annotations required to train 
a visual embedder, and to produce generalizable and semantically meaningful embeddings. For few-shot image classification we train SSL-ViTs without any supervision, on external data, and use this trained embedder to adapt quickly to novel classes with limited number of labels. For zero-shot image retrieval, we use SSL-ViTs pre-trained on a large dataset without any labels and fine-tune them with several metric learning objectives. Our self-supervised attention representations outperforms the state-of-the-art on several public benchmarks for both tasks, namely miniImageNet and CUB200 for few-shot image classification by up-to 6\%-10\%, and Stanford Online Products, Cars196 and CUB200 for zero-shot image retrieval by up-to 4\%-11\%. Code is available at \textcolor{mypink1}{\url{https://github.com/AutoVision-cloud/SSL-ViT-lowlabel-highdata}}.
\end{abstract}

\vspace{-0.1cm}

\section{Introduction}
\vspace{-0.1cm}
Self supervised representation learning has proved to be extremely successful for not only natural language processing, but also for visual recognition \cite{Simclr, mocov2}. The main reason for the growing interest in this area is the promise of overcoming the manual-annotation bottleneck while ultimately learning superior and more generalizable feature representations. Recent works like DINO \cite{DINO} have reported the emergence of semantic masks and excellent $k$-NN classifier features for self-supervised vision transformers (SSL-ViTs), outperforming its supervised counterparts on downstream tasks like image classification, copy detection and transfer learning.

\par  In the context of computer vision, especially for the low-label high-data regime, self-supervised attention
has direct applications to tasks like few-shot learning and image retrieval. The fundamental requirement for these tasks is to 
learn the semantic feature space from a small amount of labeled data, and leverage it to predict similarities between entities. The current solutions have several limitations. Best performing few-shot learning methods train models in a meta-learning fashion so it can adapt quickly on tasks with only a few training samples \cite{maml}. Meta-learning however, involves the use of complicated optimization procedures that may lead to biased representations if a large amount of labeled training examples are not available \cite{unsupmeta}. Similarly, modern methods for image retrieval focus on either learning more accurate embeddings to better reflect the semantic relations among samples, or explore more effective distance and similarity metrics \cite{musgrave}. These methods, however, also rely on the availability of a large-scale labeled dataset. Additionally, both tasks use convolutional encoders with supervised learning losses, which is prone to supervision collapse \cite{CrossTransformers}, i.e., when a neural network only represents an image's (training-set) class, and discards information that might help with generalization to out-of-distribution classes.
\vspace{0.2cm}
\begin{figure*}[ht]
    \centering
    \includegraphics[width=0.72\textwidth]{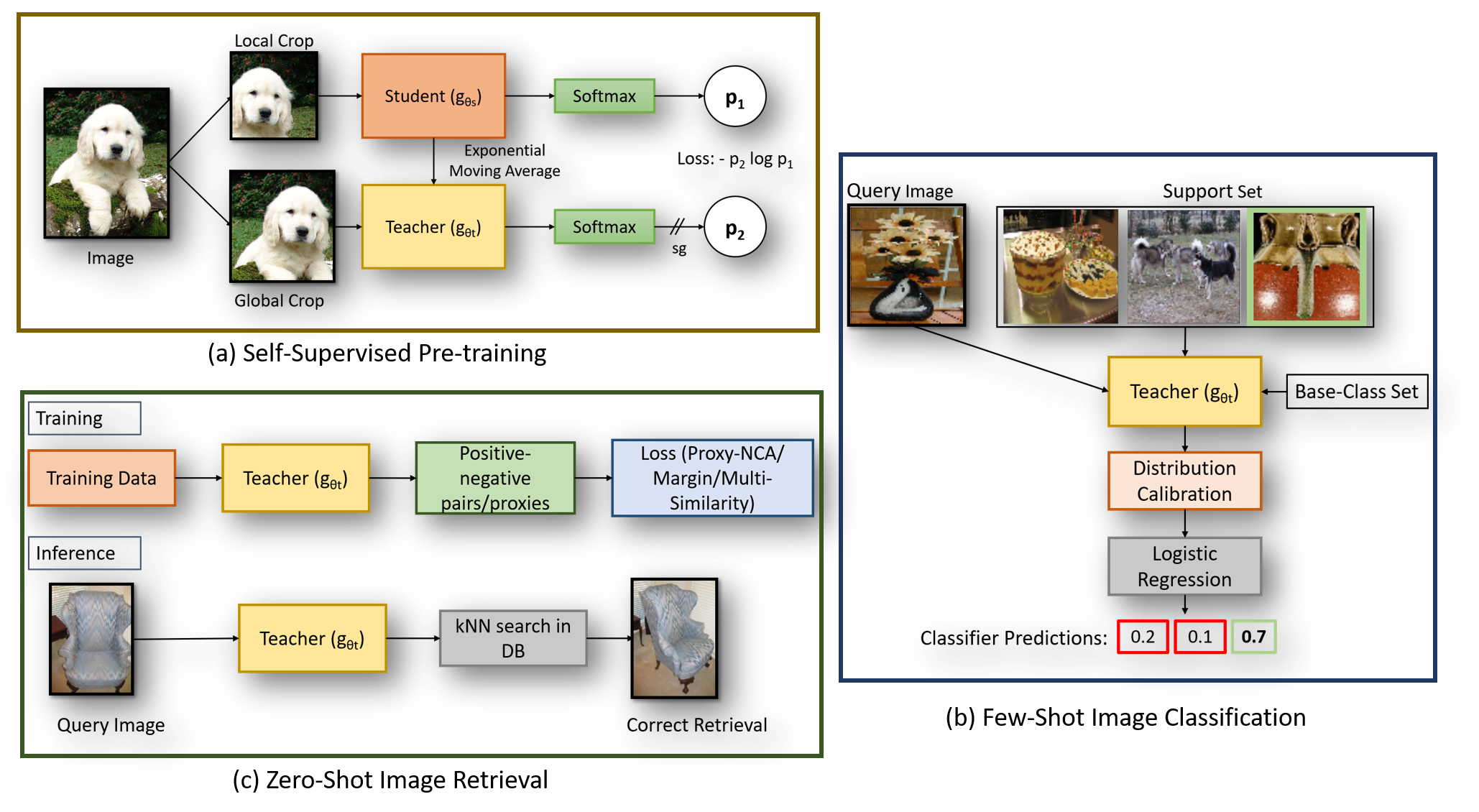}
    \caption{Proposed framework for adapting self-supervised ViT features to few-shot image classification and image retrieval.}
    \label{fig:fig1}
    \vspace{-0.4cm}
\end{figure*}
\newline 
\textbf{Contributions:} In this work, we study if the above-mentioned limitations can be solved by using self-supervised vision transformers (SSL-ViTs). We show that self-supervised attention features are extremely generalizable and can evade the annotation bottleneck. The main contributions of our work are threefold:
\vspace{-0.2cm}
\begin{itemize}[leftmargin=*, align=left]
    \item We propose a simple way to extend self-supervised vision transformers (SSL-ViTs) to two important computer vision tasks in the low-label high-data regime: few-shot image classification and zero-shot image retrieval. 
    \vspace{-0.2cm}
    \item For few-shot image classification, we show that even when SSL-ViT is trained without any supervision on external data, we can quickly adapt to novel classes with few labeled samples. Our method vastly outperforms current best-performing methods by 6\%-10\% on two public datasets. 
    \vspace{-0.2cm}
    \item For zero-shot image retrieval, we show that when SSL-ViT is pre-trained on a large dataset without any labels and subsequently fine-tuned with metric learning objectives, we can outperform convolutional architectures by 4\%-11\% on three public datasets. Interestingly, we find that we also outperform supervised ViTs by 2-7\% on two out of the three datasets.
\end{itemize}

\vspace{-0.5cm}
\section{Methods}
\vspace{-0.2cm}
Our proposed approach is illustrated in Figure \ref{fig:fig1}. We use the DINO \cite{DINO} framework to train representations from unlabelled data, with the goal of learning robust, generalizable and semantically meaningful features for few-shot learning and zero-shot image retrieval. DINO trains a student network $g_{\theta_{s}}$ to match the output of a teacher network $g_{\theta_{t}}$, parameterized by $\theta_s$ and $\theta_t$ respectively. Given an input image denoted by $\boldsymbol{x}$, output probability distribution over $K$ dimensions denoted by $P_s$ and $P_t$, and corresponding temperature parameters $\tau_s > 0$ and $\tau_t > 0$ to control the sharpness of the distribution, the probabilities are given by:
\begin{equation}
    P_s(\boldsymbol{x})^{(i)} = \frac{exp(g_{\theta_s}(\boldsymbol{x})^{(i)}/\tau_s)}{\sum_{k=1}^{K}exp(g_{\theta_s}(\boldsymbol{x})^{(k)}/\tau_s)} 
\end{equation}
In practice, from a given image, a set of $V$ different views is generated with two global views, $\boldsymbol{x}_1^g$ and $\boldsymbol{x}_2^g$ and several local views of smaller resolution. All crops are passed through the student while only the global views are passed through the teacher. Given the cross-entropy loss denoted by $H(a,b)$, the loss formulation for the self-supervised embeddings is:
\vspace{-0.05cm}
\begin{equation}
    \underset{\theta_s}{min} \sum_{\boldsymbol{x} \in \boldsymbol{x}_1^g, \boldsymbol{x}_2^g} \sum_{\underset{\boldsymbol{x}' \neq \boldsymbol{x}}{\boldsymbol{x}' \in V}} H(P_t(\boldsymbol{x}), P_s(\boldsymbol{x}')) 
\end{equation}
The parameters $\theta_s$ are learned using stochastic gradient descent. The teacher weights are frozen over an epoch and updated using a momentum encoder \cite{moco} with the update rule $\theta_t \xleftarrow[]{} \lambda \theta_t + (1-\lambda) \theta_s$, where $\lambda$ follows a cosine schedule.

\vspace{-0.2cm}
\subsection{Few-Shot Image Classification with Self-Supervised Feature Embeddings: } For our few-shot image-classification framework, we follow the distribution calibration \cite{distributioncalibration} methodology proposed as an alternative to meta-learning. The advantage is since there are no extra trainable parameters involved, we can completely side-step the complicated optimization process that characterizes the meta-learning framework. We make two key changes to the distribution calibration process. We first calculate the base-class statistics as follows:
\begin{equation}
    \boldsymbol{\mu}_i = \frac{\sum_{j=1}^{n_j}g_{\theta_t}(\boldsymbol{x}_j)}{n_j} 
    \vspace{-0.55cm}
\end{equation}
\begin{equation}
    \boldsymbol{\Sigma}_i = \frac{1}{n_j-1}\sum_{j=1}^{n_j}(g_{\theta_t}(\boldsymbol{x}_j))-\boldsymbol{\mu}_i)(g_{\theta_t}(\boldsymbol{x}_j))-\boldsymbol{\mu}_i)^T 
\end{equation}
We next calculate the support and query features for a sample $\boldsymbol{\hat{x}}$ by removing the Tukey's transformation used by the baseline \cite{distributioncalibration} and replacing it with $\boldsymbol{\Tilde{x}} = g_{\theta_t}(\boldsymbol{\hat{x}})$. We then proceed to transfer the base-class statistics to the support set using $\boldsymbol{\mu}_i$, $\boldsymbol{\Sigma}_i$ and $\boldsymbol{\Tilde{x}}$ and train a logistic regressor over the calibrated and augmented novel class distribution as described in the original formulation \cite{distributioncalibration}. 

\begin{table*}[t]
  \centering
   \begin{adjustbox}{max width=0.9\textwidth}
  \begin{tabular}{l|c|c|c|cc|cc}
    \toprule
    \multicolumn{1}{c|}{\textbf{Methods}} & 
    \multicolumn{1}{c|}{\textbf{Embedder}} & 
    
    \multicolumn{1}{c|}{\textbf{\thead{Meta-\\learning}}} &
    \multicolumn{1}{c|}{\textbf{\thead{Self-\\supervised}}} & 
    \multicolumn{2}{c|}{\textbf{miniImageNet}} &
    \multicolumn{2}{c}{\textbf{CUB}}  \\
    \cmidrule{5-6}
    \cmidrule{7-8}
     & & & & 5way1shot & 5way5shot & 5way1shot & 5way5shot \\
    
    \midrule
    MAML \cite{maml} & Conv-4 & Yes & No & 48.70 $\pm$ 1.84 & 63.10 $\pm$ 0.92 & 50.45 $\pm$ 0.97 & 59.60 $\pm$ 0.84     \\ 
    Meta-SGD \cite{meta-sgd} & Conv-4 & Yes & No & 50.47 $\pm$ 1.87 & 64.03 $\pm$ 0.94 & 53.34 $\pm$ 0.97 & 67.59 $\pm$ 0.82 \\
    Prototypical Net \cite{prototypical} &  Conv-4 & Yes & No & 54.16 $\pm$ 0.82  & 73.68 $\pm$ 0.65 & 72.99 $\pm$ 0.88 &  86.64 $\pm$ 0.51 \\
    TriNet \cite{trinet} & ResNet-18 & No & No & 58.12 $\pm$ 1.37  & 76.92 $\pm$ 0.69 & 69.61 $\pm$ 0.46 & 84.10 $\pm$ 0.35 \\
    Negative-Cosine \cite{negativecosine} & WRN-28-10 & No & No & 62.33 $\pm$ 0.82 & 80.94 $\pm$ 0.59 & 72.66 $\pm$ 0.85 & 89.40 $\pm$ 0.43 \\
    Distribution Calibration \cite{distributioncalibration} & WRN-28-10 & No & No & 68.57 $\pm$ 0.55 &   82.88 $\pm$ 0.42 & 79.56 $\pm$ 0.87 & 90.67 $\pm$ 0.35 \\
    \hline 
    Gidaris et al. \cite{gidaris} & WRN-28-10 & Yes & Yes & 63.77 $\pm$ 0.45 & 80.70 $\pm$ 0.33 & $-$  & $-$ \\
    Su et al. \cite{su} & ResNet-18 & Yes & Yes & $-$ & 76.6 $\pm$ 0.7 & $-$ & $-$ \\
    Image900-SSL \cite{amdim} & AmdimNet & Yes & Yes & 76.82 $\pm$ 0.19 & 90.98 $\pm$ 0.10 & 77.09 $\pm$ 0.21 & 89.18 $\pm$ 0.13 \\
    SSL-ViT-16 (Ours)   & ViT-S/16 & No & Yes & \textbf{86.50 $\pm$ 0.17} & \textbf{96.22 $\pm$ 0.06} & \textbf{89.94 $\pm$ 0.15} & \textbf{96.98 $\pm$ 0.05}
 \\
    \bottomrule
  \end{tabular}
  \end{adjustbox}
  \caption{5way1shot and 5way5shot classification accuracy (\%) on miniImageNet and CUB with 95\% confidence intervals.}
  \label{table1}
  
\end{table*}

\begin{table*}[t]
\centering
 \begin{adjustbox}{max width=0.7\textwidth}
\begin{tabular}{c | c |c | c | c  c c} 
 \hline
 \thead{ \textbf{Methods} \\ \\ } & \thead{\textbf{Pre-train} \\ \textbf{ImageNet}}  & \thead{\textbf{Supervised} \\ \\} & \thead{\textbf{Loss} \\ \\} & \thead{\textbf{CUB} \\ \textbf{Recall@1}} & \thead{\textbf{Cars196} \\ \textbf{Recall@1}} & \thead{\textbf{SOP} \\ \textbf{Recall@1}} \\ [0.5ex] 
 
 \hline  
 ResNet-50 \cite{musgrave} & Supervised & No & - & 48.70  & 43.50 & 52.90 \\ 
 SSL-ViT-16 (Ours) & Self-Supervised & No & - & 71.20 & 43.11 & 61.89 \\ [0.5ex] 
 \hline
 ResNet-50 \cite{diml} & Supervised & Yes & Margin \cite{margin} & 62.47 & 72.18 & 78.39 \\
 ViT-16 \cite{vit} & Supervised & Yes & Margin \cite{margin} & \textbf{81.11} & 78.95 & 80.06 \\
 SSL-ViT-16 (Ours) & Self-Supervised & Yes & Margin \cite{margin} & 74.47 & \textbf{83.59} & \textbf{81.94} \\
 [0.5ex] 
 
 \hline
 
 ResNet-50 \cite{diml} & Supervised & Yes & Proxy-NCA \cite{proxynca} & 62.76 & 71.05 & 74.70   \\
 ViT-16 \cite{vit} & Supervised & Yes & Proxy-NCA  \cite{proxynca} & \textbf{80.24} & 80.11 & \textbf{80.56} \\
 SSL-ViT-16 (Ours) & Self-Supervised & Yes & Proxy-NCA \cite{proxynca} & 71.21 & \textbf{81.67} & 75.71
 \\
 [0.5ex] 
 \hline
 
 ResNet-50 \cite{diml} & Supervised & Yes & MS \cite{multisimilarity} & 62.56 & 74.81 & 77.90   \\
 ViT-16 \cite{vit} & Supervised & Yes & MS \cite{multisimilarity} & \textbf{80.95} & 78.21 & 79.22 \\
 SSL-ViT-16 (Ours) & Self-Supervised & Yes & MS \cite{multisimilarity} & 74.40 & \textbf{85.06} & \textbf{80.36} \\ [1ex] 
 \hline
\end{tabular}
\end{adjustbox}
\caption{Recall@1 (\%) on the CUB-200-2011, Cars-196 and Stanford Online Products (SOP) datasets. }
\label{table:2}
\vspace{-0.4cm}
\end{table*}

\begin{figure*}[ht]
    \centering
    \includegraphics[width=0.74\textwidth]{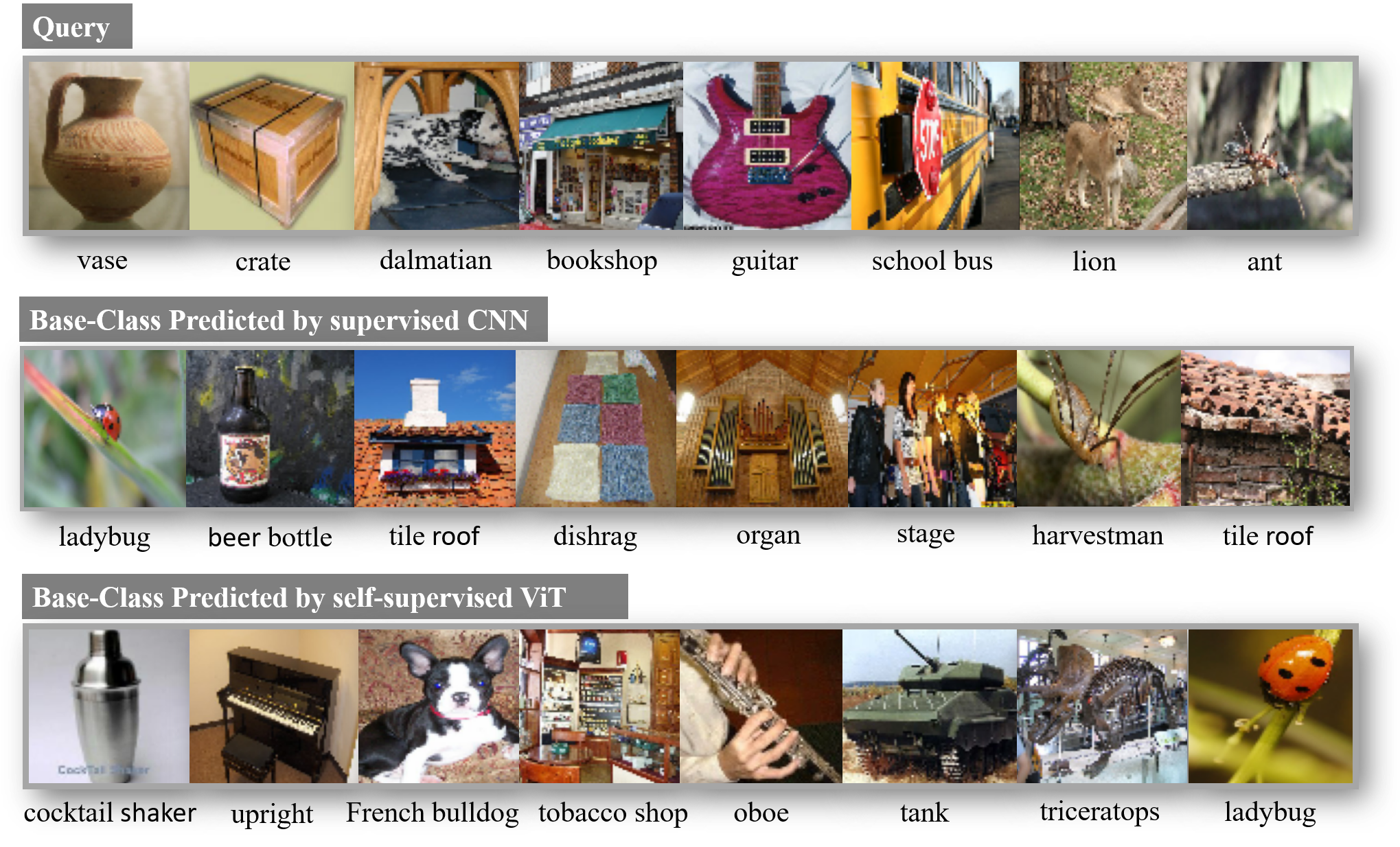}
    \caption{Self-supervised ViT chooses more semantically relevant base-classes as compared to a supervised Wide-ResNet-28 for few-shot learning with distribution calibration \cite{distributioncalibration} (for e.g., ladybug as base-class for the novel-class ant, as compared to an unrelated base-class tile-roof). The supervised convolutional counterpart can overfit to color.}
    \label{fig:comparesslvitcnn}
    \vspace{-0.4cm}
\end{figure*}
\vspace{-0.3cm}
\subsection{Zero-Shot Image Retrieval with Self-Supervised Feature Embeddings:} We follow the typical deep metric learning setup for zero-shot image retrieval \cite{musgrave}. Formally,
given a set of images $\mathcal{X} =\{\boldsymbol{x}^k\}_{k=1}^N$ and the corresponding labels $\mathcal{Y} = \{y^k\}_{k=1}^N$, deep metric learning introduces the deep neural networks $f: \mathcal{X} \xrightarrow[]{} \Phi \subset \mathbb{R}^C$ to map an image to a feature $\boldsymbol{\phi}_k = f(\boldsymbol{x}^k)$, where the semantic patterns of the input image are extracted. We make a simple change to this setup: we use the self-supervised teacher model to extract the features for an input image, given as: $\boldsymbol{\phi}_k = g_{\theta_t}(\boldsymbol{x}^k)$. We then fine-tune the embedder with several metric learning losses including Margin \cite{margin}, Proxy-NCA \cite{proxynca} and Multi-Similarity \cite{multisimilarity}.

\vspace{-0.15cm}
\section{Experimental Setup}
\vspace{-0.2cm}
\textbf{Datasets: } We evaluate SSL-ViTs on two public datasets, namely, miniImageNet \cite{miniimagenet}, and CUB-200-2011 \cite{cub} for few-shot image classification. miniImageNet contains 60,000 images from 100 classes, and each class has 600 images.  This is divided into 64 classes for training, 16 classes for validation and 20 classes for testing. CUB-200-2011(CUB) dataset contains 200 classes of birds with 11788 images. For evaluation, 200 species of birds are randomly split to 100 classes for training, 50 classes for validation, and 50 classes for testing. \\ For zero-shot image retrieval we choose three public datasets, CUB-200-2011 \cite{cub}, Cars-196 \cite{cars196} and Stanford Online Products (SOP) \cite{sop} for evaluation. For CUB, the first 100 classes (5,864 images) are used for training, and the other 100 classes (5,924 images) for testing. Cars196 contains 16,185 images of cars from 196 classes. We use the first 98 classes (8,054 images) for training and the rest 98 classes (8,131 images) for testing. SOP contains 120,053 images from 22,634 classes. We use the first 11,318 classes (59,551 images) for training and other 11,316 (60,502 images) for testing.
\vspace{0.1cm}
\newline 
\textbf{Training: }For our embedder, we use vision transformers with patch resolution of $16\times16$ \cite{vit}. Since we train with self-supervision, we name it SSL-ViT-16. The SSL pre-training occurs over 100 epochs on 4 Tesla V-100 GPUs using PyTorch 
following \cite{DINO}. For few-shot learning on mini-ImageNet, we train the embedder without any labels on 708 ImageNet classes, which are distinct from the classes present in mini-ImageNet. For few-shot learning on CUB, we train the embedder without any labels on 1K ImageNet classes which are distinct from the classes present in CUB. We do not further train on the base-classes and directly proceed to do distribution calibration, and then logistic regression on top the calibrated features for the novel classes. The embedding dimension is $384$. \\ For image-retrieval, we pre-train the SSL-ViT-16 features on 1K ImageNet classes without any labels, and then fine-tune them with three representative metric learning losses \cite{margin, proxynca, multisimilarity}. SOP and Cars-196 models
were trained for 100 epochs and the CUB-200-2011 model was trained for 20 epochs. We use embedding size
$128$ and other implementation settings following \cite{diml}. Across both tasks, we use baselines of similar capacity across a variety of architectures and supervision settings (see \cite{DINO} for capacity comparisons between ResNet-50 and SSL-ViT-16).
\vspace{0.1cm}
\newline 
\textbf{Evaluation: } For evaluation, we use the top-1 accuracy as the metric for few-shot learning, and report it on 5way-1shot and 5way-5shot settings. The reported results are the averaged classification accuracy over 10,000 tasks. For image-retrieval, we compute the Recall@1 metric.

\vspace{-0.3cm}
\section{Results}
\vspace{-0.2cm}
\par \textbf{Evaluating Few-Shot Image Classification:} 
In Table \ref{table1}, we compare our method with other meta-learning and self-supervision based methods. We show that SSL-ViT-16 outperforms all baselines across both miniImageNet and CUB for both 5way-1shot and 5way-5shot settings.  
Our method also closes the gap between 5way-1shot with the 5way-5shot setting by reducing the difference from 14.16\% to 9.72\% for mini-ImageNet and from 12.09\% to 7.04\% 
for CUB. This indicates that our visual embedder has extracted generalized, robust and high quality features. SSL-ViT-16 also convincingly outperforms the distribution calibration baseline \cite{distributioncalibration} by 10\%-18\% for the more challenging 5-way 1-shot setting with a model of similar capacity. 
Note that ViT-16 has 21M parameters compared to WRN-28-10 which has 36M. We also outperform other recently proposed self-supervised approaches including auxiliary training setups \cite{gidaris, su} by up-to 22\%, as well as the Image900-SSL \cite{amdim} which was pre-trained on 900 ImageNet classes (as opposed to ours which was pre-trained on 708 classes) by up-to 10\%. This shows the power of attention features when trained with self-supervision. In Figure \ref{fig:comparesslvitcnn}, we also qualitatively show that SSL-ViT identifies more semantically relevant base-classes for distribution calibration compared to supervised convolution, even though we trained it without labels.
\vspace{0.1cm}
\newline 
\textbf{Evaluating Zero-Shot Image Retrieval:} In Table \ref{table:2}, we compare our method with a supervised ResNet-50 and a supervised ViT-16 baseline. We show that SSL-ViT-16 outperforms convolutional baselines \cite{diml} by up-to 4\%-11\% across all three datasets and metric losses. Interestingly we also find SSL-ViT-16 outperforms supervised ViT-16 on the comparatively more challenging Cars-196 and SOP datasets across Margin and Multi-Similarity losses. This suggests that while supervised transformers are attractive metric learners, self-supervision can lead to further gains possibly due to its ability to infer semantic layouts. This establishes them as an attractive alternative to convolutional methods for image retrieval. The \textit{main advantage} however is that we can use large unlabeled datasets for pre-training to create stronger initial features. From this section, we thus infer that self-supervision with attention produces robust, transferable and superior features as compared to directly supervised convolutional and attention features.

\vspace{-0.35cm}
\section{Conclusion}
\vspace{-0.3cm}
In this paper, we successfully adapt self-supervised attention to two important computer vision tasks in the low-label high-data regime: few-shot learning and zero-shot image retrieval. 
We show that SSL-ViTs capture better semantic meanings from completely unlabeled data and also improve state-of-the-art performance across five public datasets.
Our experiments provide insights into the data-efficiency and generalizability of self-supervised attention representations and can be used to advance future research in this direction.

\bibliographystyle{IEEEbib}
{\footnotesize
\bibliography{refs.bib}}
\end{document}